\documentclass[letterpaper, 10 pt, conference]{ieeeconf}  
\IEEEoverridecommandlockouts                              
\usepackage{graphicx}      

\usepackage{xcolor}
\usepackage{mathtools}

\usepackage{epsfig} 
\usepackage{mathptmx} 
\usepackage{times} 
\usepackage{amsmath} 
\usepackage{amssymb}  
\usepackage{leftidx}

\newtheorem{lemma}{Lemma}
\newtheorem{definition}{Definition}

\newtheorem{proposition}{Proposition}
\newtheorem{assum}{Assumption}
\newtheorem{remark}{Remark}
\usepackage{color}
\DeclareSymbolFont{newfont}{OML}{cmm}{m}{it}
\DeclareMathSymbol{\Varrho}{3}{newfont}{37}

\usepackage{algorithm,algorithmic}
\usepackage{bm}
\usepackage{hyperref}

\DeclareMathOperator*{\argmin}{arg\,min}

\title{\LARGE \bf
Collision-free landing of multiple UAVs on moving ground vehicles using time-varying control barrier functions
}
\author{Viswa Narayanan Sankaranarayanan$^*$, Akshit Saradagi, Sumeet Satpute and George Nikolakopoulos
\thanks{$^*$ Corresponding author}
\thanks{The authors are with the Robotics and Artificial Intelligence Group at the Department of Computer Science, Electrical and Space Engineering, Lule\aa~University of Technology, Sweden. Emails: \tt{vissan@ltu.se; akssar@ltu.se; sumsat@ltu.se; geonik@ltu.se}}%
}

\begin{document}

\maketitle
\thispagestyle{empty}
\pagestyle{empty}

\begin{abstract}
In this article, we present a centralized approach for the control of multiple unmanned aerial vehicles (UAVs) for landing on moving unmanned ground vehicles (UGVs) using control barrier functions (CBFs). The proposed control framework employs two kinds of CBFs to impose safety constraints on the UAVs' motion. The first class of CBFs (LCBF) is a three-dimensional exponentially decaying function centered above the landing platform, designed to safely and precisely land UAVs on the UGVs. The second set is a spherical CBF (SCBF), defined between every pair of UAVs, which avoids collisions between them. The LCBF is time-varying and adapts to the motions of the UGVs. In the proposed CBF approach, the control input from the UAV's nominal tracking controller designed to reach the landing platform is filtered to choose a minimally-deviating control input that ensures safety (as defined by the CBFs). As the control inputs of every UAV are shared in establishing multiple CBF constraints, we prove that the control inputs are shared without conflict in rendering the safe sets forward invariant. The performance of the control framework is validated through a simulated scenario involving three UAVs landing on three moving targets.
\end{abstract}

\section{Introduction} \label{sec:intro}
Coordinated multi-agent systems with unmanned ground vehicles (UGVs) and unmanned aerial vehicles (UAVs) have become popular in numerous applications \cite{lindqvist2022multimodality, pretto2020building}. UAVs and UGVs complement each other by expanding the scope of the multi-agent system by introducing additional degrees of freedom and optimizing fuel consumption in the overall mission, especially in inspection and surveying tasks \cite{cantelli2013uav}. A vital goal in these missions is landing the UAVs on their respective UGVs. The landing operation introduces several challenges to the system, mainly when the UGVs are mobile \cite{persson2017cooperative}. In the literature, the landing problem is addressed to primarily focus on customizing the design of the landing platform for an efficient landing, detecting the landing region based on visual cues, and tracking the position of a landing platform \cite{grlj2022decade}.

In practice, the landing operation is carried out in a multi-stage switching operation to safeguard the UAV from colliding with the landing platform and to avoid aerodynamical effects when it flies close to the platform. The switching sequence involves three phases: rising, approaching, and descending. In the rising phase, the UAV rises to a convenient altitude to then move towards the target without colliding with the platform. In the approach phase, the UAV moves horizontally to align itself over the platform vertically, whereas, in the descending phase, the UAV descends vertically to land on the platform.

The computer vision and trajectory tracking-based landing solutions assume the UAV to be at a convenient altitude and ignore the aerodynamic effects while flying close to the surface (cf. \cite{lee2016vision, zou2020robust}). Moreover, the precise landing of the UAV on a moving UGV is addressed as a prescribed performance tracking control problem or an accurate estimation problem (cf. \cite{kalinov2019high, ganguly2021efficient, ganguly2022robust, sankaranarayanan2022robustifying, sankaranarayanan2021adaptive}). In both cases, the controllers are implemented as a setpoint tracking problem to approach the landing position. However, a systematic control strategy to shape the landing maneuver for the UAV to land on the moving UGV while ensuring the safety of both the robots needs to be better explored. Further, all of the works in the literature consider the landing problem of a single UAV on a UGV. In contrast, this paper presents a control strategy for designing safe maneuvers for multiple UAVs to land on their corresponding moving UGVs.

\begin{figure}[!h]
	\centering
	\includegraphics[width=0.48\textwidth]{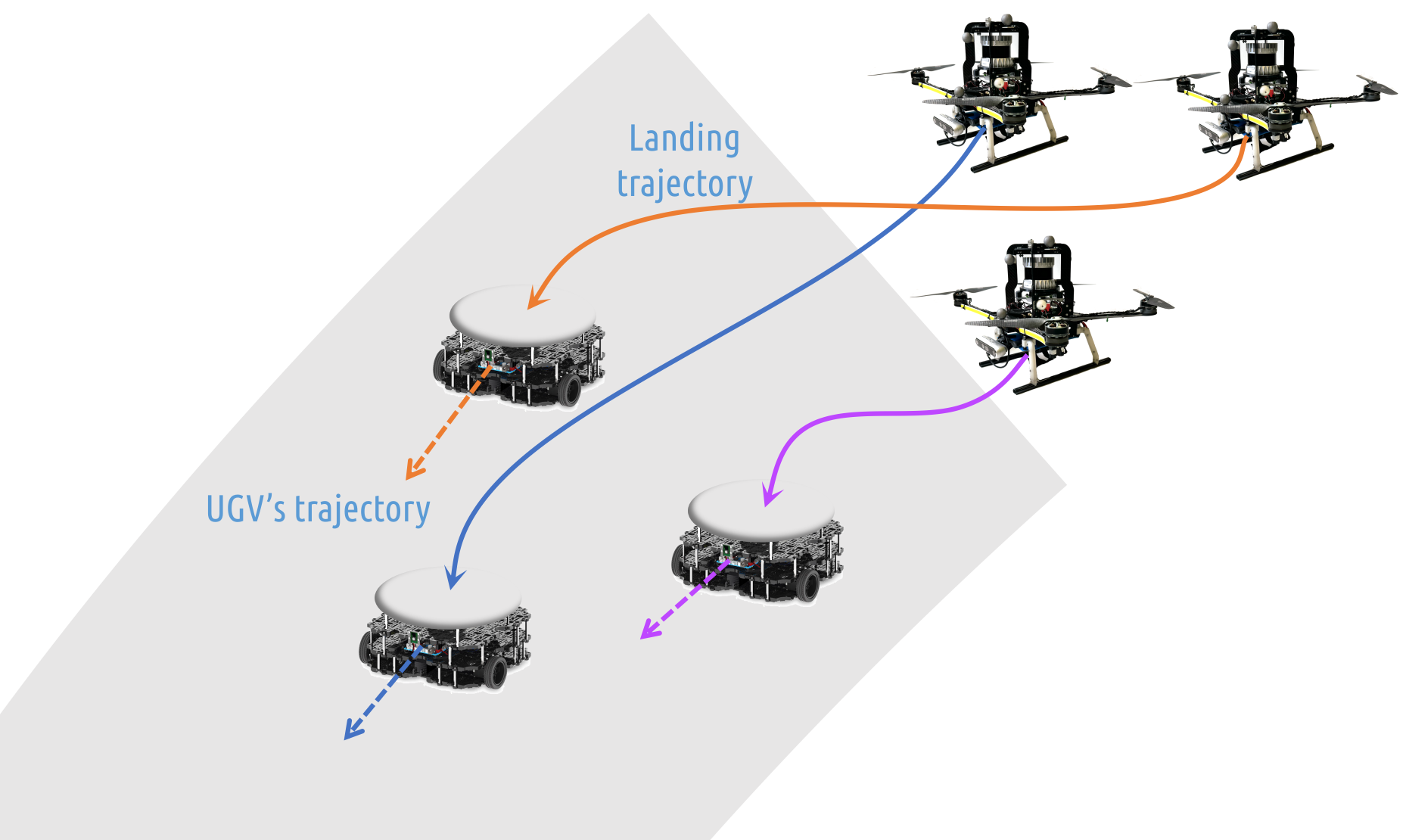}
	\caption{{A schematic of multiple UAVs landing on their respective UGVs.}}\label{fig:uav_model}
\end{figure}
Along with generating a safe landing maneuver to avoid crashing into the UGVs, a multi-UAV landing problem also introduces the challenges of preventing the collision between the UAVs. Therefore, shaping the landing maneuver becomes a control problem involving multiple nonlinear constraints. While model predictive controllers are used in constrained optimization problems \cite{lindqvist2020nonlinear}, the nonlinearities in the constraints introduce infeasibilities and increase computational complexity.

Control barrier functions (CBFs) have evolved to be an elegant way to enforce safety guarantees. In this approach, a safe set, defined as the super level set of the control barrier function, is rendered forward invariant and asymptotically stable \cite{ames2019control}. The CBFs convert the nonlinear constraints in the state space into a set of affine constraints in the control space, thus minimally deviating from the nominal control inputs while enforcing safety in the closed-loop operation. CBFs are widely used for various robotic problems, such as collision avoidance \cite{qing2021collision}, space docking \cite{saradagi2022safe}, terrain navigation \cite{nguyen20163d}, aerial manipulation \cite{berra2024assisted} etc. Recently, CBFs have made their way into control of UAVs, especially in visual surveying and collision avoidance applications. Moreover, in \cite{zhou2020control}, the bounded error position tracking problem for UAV landing has also been addressed using CBF approaches. However, a CBF-based approach for multi-UAV landing with guaranteed safety is still missing in the literature.

\textbf{Contributions:} Given this premise, the contributions of this article are summarized as follows: (a) A CBF-based centralized control approach is proposed that enables multiple UAVs to land on moving UGVs safely and precisely. 
Unlike \cite{lee2016vision, zou2020robust, kalinov2019high, guo2020precision}, the proposed method considers a systematic maneuver of multi-UAVs to handle all three phases of rising, approaching, and descending. (b) The LCBF forms a single time-varying constraint centered at the position of each of the UGVs to handle all three phases involved in landing on moving targets. Further, the parameters that shape the landing maneuver are parameterizable based on the dimensions of the robots and available safety margins. (c) An additional safety layer to avoid collisions between the UAVs is provided using SCBFs, which are defined for every pair of UAVs. (d) The theoretical proof of harmonious control input sharing between the CBFs ensuring the forward invariance of the safety sets is presented. 

\section{Notations and Preliminaries} \label{sec:not_pre}
The notations $\mathbb{R}$ and $\mathbb{S}^1$ denote real numbers and unit circles respectively; $\mathbf{I}$ denotes identity matrix of appropriate dimensions; the variable $i$ used in the subscript refers to the attribute of the $i^{th}$ UAV for all $i \in N$ unless mentioned otherwise; the variable $j$ in the subscript refers to the attribute of the $i^{th}$ UAV corresponding to $j^{th}$ UAV for all $\lbrace j \in N ~|~ j > i \rbrace$; 
$\partial \mathcal{B}$ denotes the boundary of the closed set $\mathcal{B}$; $L_{f}h(p) \triangleq \frac{\partial h(p)}{\partial p} f(p)$ denotes the Lie derivative of a function $h(p)$ along $f(p)$.
\begin{definition}[Class-$\mathcal{K}$ function]
    A continuous function, $\omega:(-b,a) \xrightarrow{}(-\infty, \infty)$ is an extended class-$\mathcal{K}$ function if the function is strictly increasing and $\omega(0) = 0$. Moreover, the function is an extended class-$\mathcal{K}_\infty$ function if $(-b,a)=(-\infty, \infty)$.
\end{definition}

Further, let us state the fundamentals of the control barrier function, which would be used as a filter to ensure safety constraints during the UAVs' landing maneuvers. Let the time-varying safe set for a given affine dynamical system, $\dot{p}(t) = f(p(t)) + g(p(t))u(t)$ be defined by the state space region, $\mathcal{S}(t) \subset \mathcal{P}$. Now, the continuously differentiable function $h(p,t): \mathcal{D}(t) \subset \mathcal{P} \xrightarrow{} \mathbb{R}$, such that $\mathcal{S}(t):\lbrace p \in \mathcal{P} | h(p,t) \geq 0 \rbrace$ ($\mathcal{S}(t)$ is a zero-level super set of $h(p,t) \forall t$), renders the set $\mathcal{S}(t)$ safe if the control input to $u$ ensures positive invariance of the set $\mathcal{S}(t) ~ \forall t$, i.e.,  $p(t_0) \in \mathcal{S}(t_0)$ implies $p(t) \in \mathcal{S}(t) ~ \forall t \geq t_0$. Moreover, if the safe set, $\mathcal{S}(t)$ is rendered asymptotically stable when $p(t)$ is initialized outside the unsafe set ($p(0) \in \mathcal{D}(0) \setminus \mathcal{S}(0)$), a measure of robustness can be incorporated into the notion of safety. The condition validating the continuous function, $h(p,t)$ as a control barrier function, is presented in the following definition.
\begin{definition}[Time-varying CBF] \label{Def:TV-CBF}
    A candidate function $h(p,t)$ is a valid control barrier function for an affine dynamical system, $\dot{p} = f(p,t) + g(p,t)u$ if there exists a locally Lipschitz continuous class-$\mathcal{K}_\infty$ function $\omega$, such that $\forall (p,t) \in \mathcal{D}(t) \times [0, \infty]$
    \begin{align}
        \sup_{u \in \mathcal{U}} \frac{\partial h(p,t)^T}{\partial p} \left (f(p) + g(p)u  \right ) + \frac{\partial h(p,t)}{\partial t} \geq -\omega(h(p,t)) \label{eq:cbf_def}
    \end{align}
\end{definition}
\begin{proposition} \label{prpsn:asymptotic}
The zero super-level set $\mathcal{S}(t)$ of the continuously differentiable function $h(p,t): \mathcal{D}(t) \subset \mathcal{P} \xrightarrow{} \mathbb{R}$ is rendered forward invariant and asymptotically stable in $\mathcal{D}(t)$ by a Lipschitz continuous controller $u(p,t) \in \kappa (p,t)$, where
\begin{align}
    \kappa(p,t) = & \lbrace u \in \mathcal{U} | \frac{\partial h(p,t)^T}{\partial p} \left (f(p) + g(p)u  \right ) + \frac{\partial h(p,t)}{\partial t} \nonumber \\
    & + \omega(h(p,t)) \geq 0 \rbrace \label{eq:con_inp}
\end{align}
for the system $\dot{p} = f(p,t) + g(p,t)u$, if $h(p,t)$ is a valid control barrier function on $\mathcal{D}(t)$ and $\frac{\partial h(p,t)}{\partial p} \neq 0$ on $\partial \mathcal{S}$.
\begin{remark}
The condition in \eqref{eq:cbf_def} captures the forward invariance of $\mathcal{S}(t) ~(\dot{h}(p,t) \geq 0 \text{ on } \partial \mathcal{B}(t))$, and the asymptotic stability of $\mathcal{S}(t) ~ (\dot{h}(p,t) \geq 0 \text{ in } \mathcal{D}(t)\setminus \mathcal{S}(t))$. However, in this article, we utilize only the forward invariance property in the design of CBF-based safety filters for safe multi-UAV landing.
\end{remark}
%
%
\end{proposition}

\section{Problem Formulation} \label{sec:prob_form}
In this section, the UAV is modeled using its dynamics, and the constraints required for the safe and interference-free landing of multiple UAVs on their respective moving UGVs are identified.

\subsection{Modeling of the UAV} \label{subsec:model_uav}
\begin{figure}[!h]
	\centering
	\includegraphics[width=0.3\textwidth]{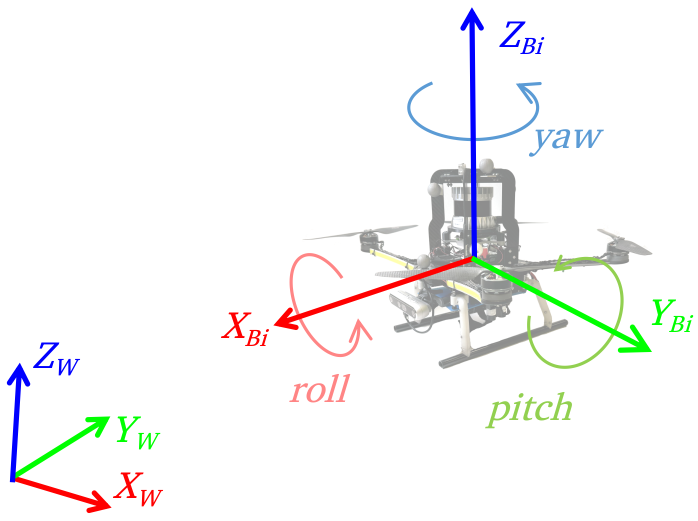}
	\caption{{A schematic of a quadrotor UAV with the associated coordinate frames.}}\label{fig:uav_model}
\end{figure}
The dynamics of a multi-rotor UAV is modeled as,
\begin{subequations} \label{eq:sys_model}
\begin{align}
    m_i {\ddot{p}_i}(t) + m_i {G} &= \tau_{pi}, \label{eq:p_tau} \\
    {J_i}(q_i(t)) {\ddot{q}_i}(t) + {C_i}({q_i, \dot{q}_i, t}){\dot{q}_i} (t) &= {\tau}_{{qi}}, \label{eq:q_tau} \\
		\tau_{pi} &= {R_i} {F_i} \label{eq:force_map}
\end{align}    
\end{subequations}

where $p_i(t) \triangleq \begin{bmatrix}
    x_i(t), y_i(t), z_i(t)
\end{bmatrix}^T \in \mathbb{R}^3$, $q_i(t) \triangleq \begin{bmatrix}
    \phi_i(t), \theta_i(t), \psi_i(t)
\end{bmatrix}^T$ represents the position and orientation of the UAV in the inertial frame ($\mathbf{X_W - Y_W - Z_W}$); $G \triangleq \begin{bmatrix}
    0, 0, -g
\end{bmatrix}^T \in \mathbb{R}^3$ is the gravity vector with $g = 9.81 ms^{-2}$ is the acceleration due to gravity; $\tau_{pi} \triangleq \begin{bmatrix} \tau_{xi}, \tau_{yi}, \tau_{zi} \end{bmatrix}^T  \in \mathbb{R}^3$ is the position control inputs in the inertial frame, which is mapped to the body-frame ($\mathbf{X_{Bi}-Y_{Bi}-Z_{Bi}}$, cf. Fig. \ref{fig:uav_model}) control input $F_i \triangleq \begin{bmatrix}
    f_{xi} & f_{yi} & f_{zi}
\end{bmatrix}^T \in \mathbb{R}^3$ using the rotation matrix, $R_i \in \mathbb{R}^{3 \times 3}$; $m_i$ is the mass and $J_i, C_i \in \mathbb{R}^{3 \times 3}$ are the inertia and Corilosis matrices of the UAV; $\tau_{qi} \in \mathbb{R}^3$  is the attitude control input of the $i^{th}$ UAV for $i \in \lbrace 1, \cdots, N \rbrace $ and $N$ is the total number of UAVs.

\subsection{Constraints for Safe Landing} \label{subsec:landing_cons}
The landing task of a UAV over a moving platform is typically performed in a sequence of three subtasks, namely:
\begin{enumerate}
    \item Altitude adjustment to stay conveniently above the altitude of the landing region.
    \item Horizontal approach to align itself along the Z-axis of the UGV.
    \item Vertical descent towards the landing platform.
\end{enumerate}
This switching sequence is required to ensure the essential safety constraints of collision avoidance between the UAV and the UGV landing platform, reduction of aerodynamic effects by vertical descent, and clearance from the ground are maintained in the landing process. Further, the following assumption is made on the velocity of UGV for performing the landing maneuver.

\begin{assum}[Velocity of the UGV] \label{as:velocity}
    The velocities of the UGVs $\dot{p}_{di}(t) \triangleq
	\begin{bmatrix}
	\dot{x}_{di} (t) & \dot{y}_{di} (t) & \dot{z}_{di} (t)
	\end{bmatrix}^T $ are smooth and bounded, such that $||\dot{p}_{di}(t)|| < \sigma \; \forall \,t \geq 0$ and $\sigma$ is a positive real number.
\end{assum}

\begin{figure}[!h]
	\centering
	\includegraphics[width=0.45\textwidth]{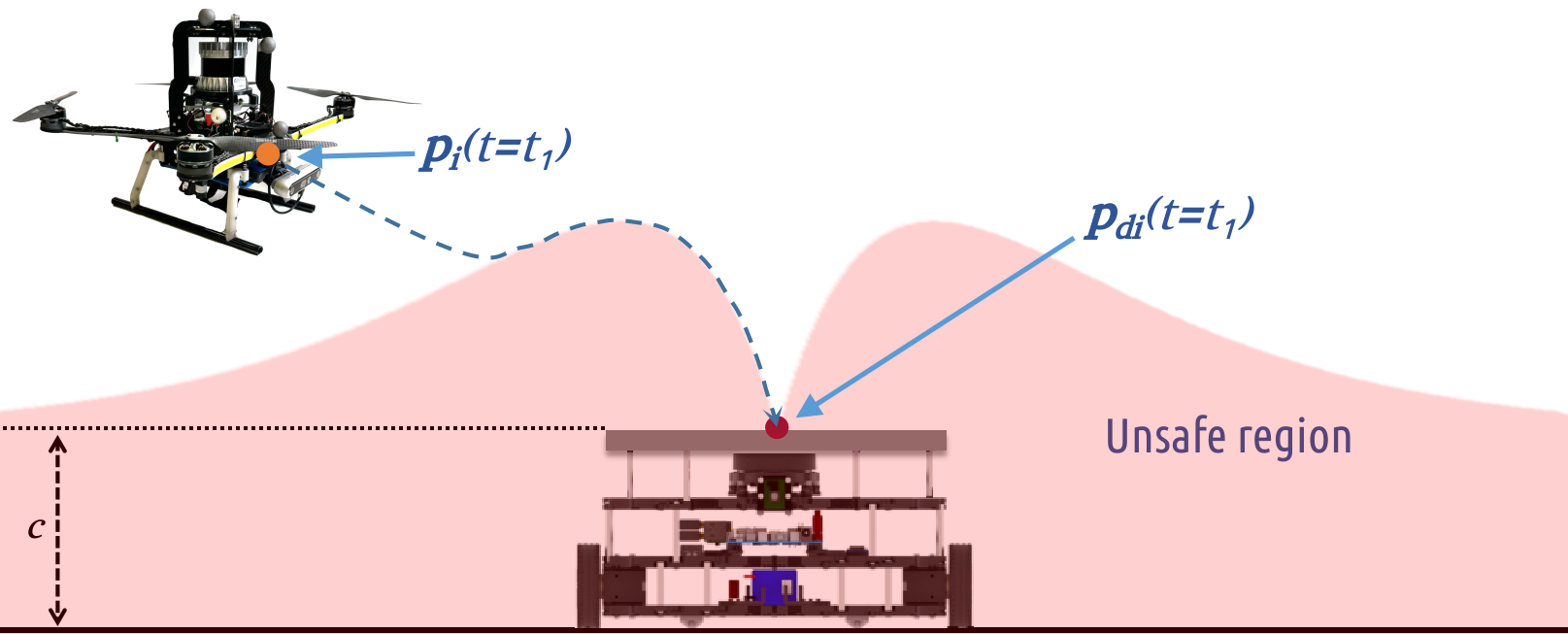}
	\caption{{The constraint for landing a UAV on a UGV combines the properties of ground clearance $c$ and approaching the UGV only from a conical region  above the landing position $p_{di}(t=t_1)$. The UAV must perform the landing from its initial position $p_{i}(t=t_1)$ without entering the unsafe region.}}\label{fig:lcbf_constraint}
\end{figure}

For constructing a CBF just to fulfill a simple landing problem for a UAV\textendash UGV pair, it is necessary to identify the safe and unsafe regions defined by these constraints. For the collision avoidance constraint between the UAV and the UGV and to provide ground clearance, the region in space below the UAV's altitude is deemed unsafe. Such a constraint prevents the UAV from colliding with the UGV while still allowing it to approach the UGV from the top. The vertical descending in the proximity of the UGV is imposed by defining a conical region of safety on top of the UGV, such that the region outside the cone is deemed unsafe for descending. The switched sequence operation of landing could be confined to a single landing maneuver by designing the constraints for the landing CBF (LCBF) to be,
\begin{align}
    e_{zi}(t) &>=  \beta_i \alpha_i d_i(t)\exp(- \alpha_i d_i(t)), \label{eq:lcbf_constraint} \\
    d_i(t) &=  \sqrt{e_{xi}^2(t) + e_{yi}^2(t)}, \label{eq:d}
\end{align}
where $\alpha_i, \beta_i$ are user-defined scalars for shaping the LCBF, and $e_{xi}(t), e_{yi}(t), e_{zi}(t)$ form the position error vector,
\begin{align}
    e_{pi}(t) &= 
    \begin{bmatrix}
        e_{xi}(t) \\ e_{yi}(t) \\ e_{zi}(t)
    \end{bmatrix} = \begin{bmatrix}
       x_i(t) - x_{di}(t) \\ y_i(t) - y_{di}(t) \\ z_i(t) - z_{di}(t)
    \end{bmatrix}. \label{eq:err_vec}
\end{align}
where $p_{di}(t) \triangleq \begin{bmatrix}
    x_{di}(t), y_{di}(t), z_{di}(t)
\end{bmatrix}$ is the $i^{th}$ UAV's landing spot on its corresponding UGV. The unsafe region created by this constraint is illustrated in Fig. \ref{fig:lcbf_constraint}. 
\begin{figure}[!h]
	\centering
	\includegraphics[width=0.45\textwidth]{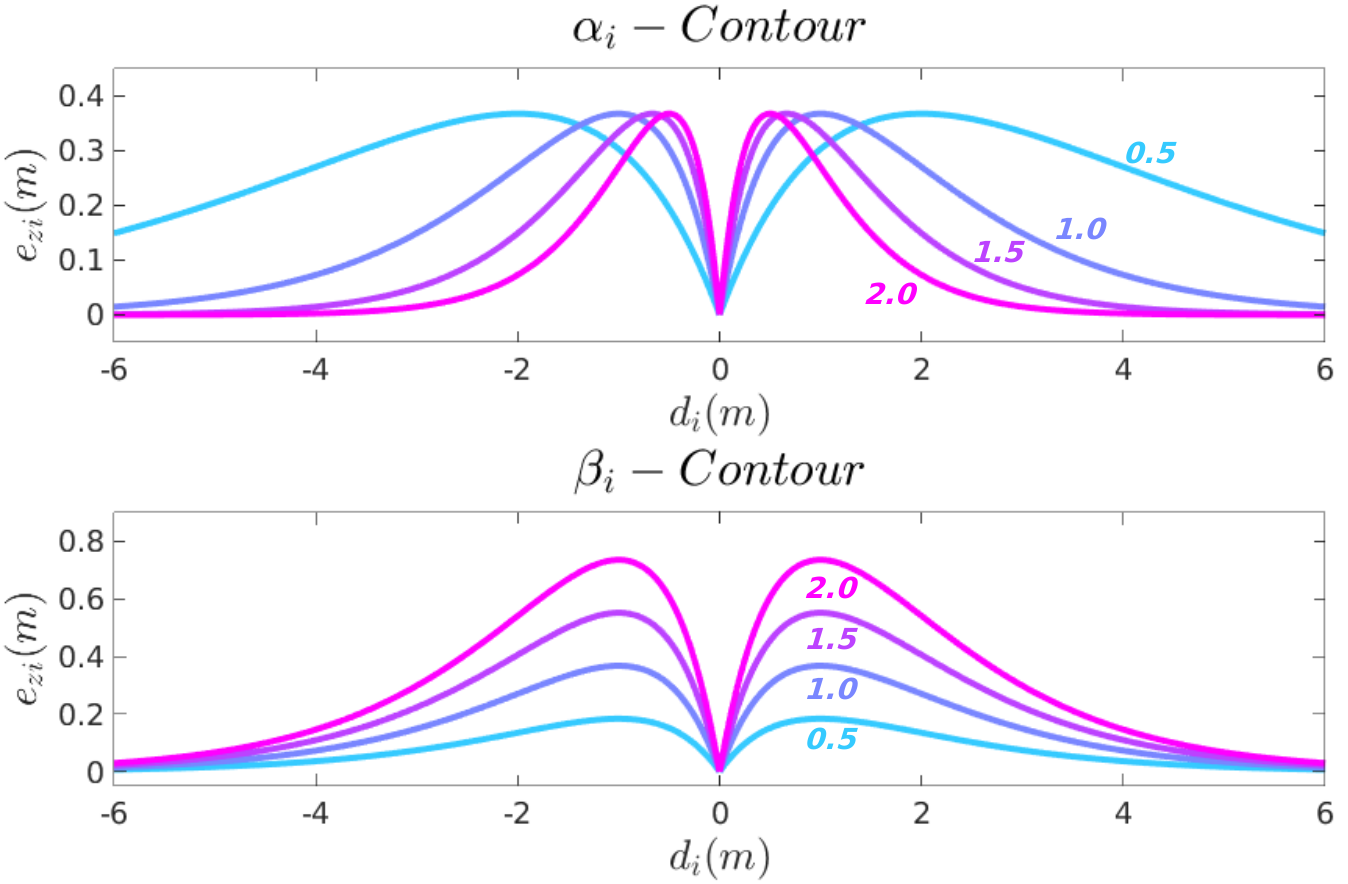}
	\caption{{$\alpha_i$ parameters scale the safe regions horizontally, whereas $\beta_i$ parameters scale them vertically.}}\label{fig:ab_contour}
\end{figure}

\begin{remark}
    Along with the controlled states (position of the UAV), the landing constraint in \eqref{eq:lcbf_constraint} is also dependent on the external state (UGV's position), which varies with time, making the constraint time-varying.
\end{remark}

The parameters $\alpha_i, \beta_i$ scale the boundary layer of the landing constraint horizontally and vertically, respectively (cf. Fig. \ref{fig:ab_contour}). Their values are chosen by identifying the peak altitude of the boundary by finding its partial derivative with respect to the horizontal distance, $d_i(t)$.
\begin{align}
    \frac{\partial e_{zi}(t)}{\partial d_i (t)} &= \beta_i \alpha_i (1 - \alpha_i d_i) \exp (-\alpha_i d_i), \\
    \frac{\partial e_{zi}(t)}{\partial d_i (t)} &= 0 \implies \alpha_i = \frac{1}{d_i} \implies \beta_i = e^*_{zi} \exp(1), \label{eq:ab_values}
\end{align}
where $e^*_{zi}$ is the peak altitude of the boundary layer and $\exp(1) \approx 2.718$. Using \eqref{eq:ab_values}, $\alpha_i, \beta_i$ are chosen by deciding the radius and altitude ($d_i, e^*_{zi}$) of the conical descending region.

\subsection{Constraints for Collision Avoidance} \label{subsec:collision_cons}
In the case of multiple UAV\textendash UGV system, it is inevitable to design a safe maneuver without constraining the UAVs to avoid colliding with each other. The safe and unsafe regions are classified in this case by bounding the UAVs with a sphere of radius, $s_i$. The following assumption highlights the necessary condition for formulating the collision avoidance constraint. Therefore, the SCBF constraint between the $i^{th}$ UAV and $j^{th}$ UAV is given by,
\begin{align}
    ||p_i(t) - p_j(t)||^2 > (s_i + s_j)^2. \label{eq:scbf_constraint}
\end{align}

\begin{assum} [Distance between UGVs] \label{as:position}
    The minimum distance between any two UGVs ($i,j$) is always greater than the sum of the bounding radii of all pairs of UAVs, such that $||p_{di}(t) - p_{dj}(t)|| > \text{max} (s_m + s_n), ~ \forall 1 < m < n < N$.
\end{assum}

\noindent\textbf{Problem Statement:} Design a centralized control architecture for the multi-UAV system for landing on their corresponding moving UGVs enforcing the constraints stated in eqs. \eqref{eq:lcbf_constraint} \& \eqref{eq:scbf_constraint} under the assumptions \ref{as:velocity} \& \ref{as:position}, given the position and velocity of the UGVs and the positive real design parameters $\alpha_i, \beta_i, s_i, a_i, b_i, m_i, \forall \;i \in \lbrace 1,...,N \rbrace$.

\section{Control Architecture} \label{sec:con_arch}
\begin{figure}[!h]
	\centering
	\includegraphics[width=0.48\textwidth]{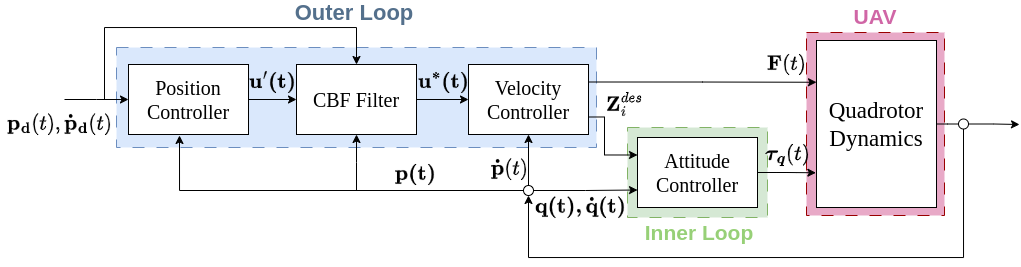}
	\caption{{Block diagram of the proposed control architecture.}}\label{fig:arch}
\end{figure}

The control of the UAVs is carried out using a centralized architecture, with a dual loop control for the position and attitude of the UAV (cf. Fig. \ref{fig:arch}). The CBF filter is implemented inside the outer loop. The following subsections detail the position and attitude control loops and the CBF filter.
\subsection{Position Control Loop}
The position control loop is divided into three blocks: (a) the position tracking block, (b) the CBF block acting as a velocity input filter, and (c) the velocity tracking block. Let $\mathbf{p} \triangleq \begin{bmatrix}
    p_1^T & p_2^T & \cdots & p_N^T
\end{bmatrix}^T,~ \mathbf{p_d} \triangleq \begin{bmatrix}
    p_{d1}^T & p_{d2}^T & \cdots & p_{dN}^T
\end{bmatrix}^T \in \mathbb{R}^{3N}$ be the appended position vector of all the UAVs and their corresponding landing positions in the inertial frame. A pseudo-velocity input is generated using the following equation,
\begin{align}
    \mathbf{u}'(\mathbf{p},t) &= \mathbf{K_p}(\mathbf{p_d}(t) - \mathbf{p}(t)) + \mathbf{\dot{p}_d}(t) \label{eq:position_control}
\end{align}
where $\mathbf{K_p} \in \mathbb{R}^{3N \times 3N}$ is a positive definite gain matrix. It is to be noted that the UGV's velocity (desired velocity, $\mathbf{\dot{p}_d}$) is added in the eq. \eqref{eq:position_control} so that the combined velocity input (pseudo-velocity input, $\mathbf{u}'(\mathbf{p},t) \in \mathbb{U} = \mathbb{R}^{3N}$) would be filtered using the CBF block to obtain the actual velocity input, $\mathbf{u^*}$ to be given as an input to the velocity tracking block. The CBF block is explained further in a later section \ref{sec:cbf_filter}. The velocity control input, $\mathbf{u^*}$ is tracked using the velocity tracking block as given below,
\begin{align}
    \boldsymbol{\tau}_{\mathbf{p}} &= \mathbf{K_v}(\mathbf{u^*} - \mathbf{\dot{p}}) + \mathbf{M}\mathbf{G}, \label{eq:velocity_control} \\
    \mathbf{M} &= \begin{bmatrix}
        m_1 & 0 & 0 & 0 & \cdots & 0 \\
        0 & m_1 & 0 & 0 & \cdots & 0 \\
        0 & 0 & m_1 & 0 & \cdots & 0 \\
        0 & 0 & 0 & m_2 & \cdots & 0 \\
        . & . & . & . & \cdots & . \\
        0 & 0 & 0 & 0 & \cdots & m_N
    \end{bmatrix}, \nonumber
\end{align}
where $\boldsymbol{\tau}_{\mathbf{p}} \triangleq \begin{bmatrix}
    \tau_{p1}^T & \tau_{p2}^T & \cdots & \tau_{pN}^T
\end{bmatrix}^T \in \mathbb{R}^{3N}$ is the appended position control vector in the inertial frame, $\mathbf{K_v} \in \mathbb{R}^{3N \times 3N}$ is a positive definite gain matrix, $\mathbf{G} \triangleq \begin{bmatrix}
    G^T & G^T & \cdots & G^T
\end{bmatrix}^T \in \mathbb{R}^{3N}$. The control inputs in \eqref{eq:velocity_control} is converted to body-frame position control using the following equation,
\begin{align}
    \mathbf{F} &= \mathbf{R}\boldsymbol{\tau}_{\mathbf{p}}, \label{eq:thrust_input} &
    \mathbf{R} = \begin{bmatrix}
        R_1 & \mathbf{0} & \cdots & \mathbf{0} \\
        \mathbf{0} & R_2 & \cdots & \mathbf{0} \\
        . & . & \cdots & . \\
        \mathbf{0} & \mathbf{0} & \cdots & R_N \nonumber
    \end{bmatrix},
\end{align}
where $\mathbf{F} \triangleq \begin{bmatrix}
    F_1^T & F_2^T & \cdots & F_N^T
\end{bmatrix} \in \mathbb{R}^{3N}$ is the appended force vector in the body-frame, $\mathbf{0} \in \mathbb{R}^{3 \times 3}$ is a square matrix with all its elements as zeroes. The thrust vector for the UAV is given by, $\mathbf{F_z} \triangleq \begin{bmatrix}
    f_{z1} & f_{z2} & \cdots & f_{zN}
\end{bmatrix} \in \mathbb{R}^N$.

\subsection{Attitude Control Input} \label{subsec:att_control}
We utilize the geometric control approach for the attitude control loop as presented in \cite{lee2010geometric}. The desired body frame is obtained by,
\begin{subequations}
    \begin{align}
        Z_{Bi} &= \tau_{pi}/||\tau_{pi}||, ~
        Y_{Ai} = \begin{bmatrix}
            -\sin(\psi_{di}) & \cos(\psi_{di}) & 0
        \end{bmatrix}^T, \\
        X_{Bi} &=  \frac{Y_{Ai} \times Z_{Bi}}{||Y_{Ai} \times Z_{Bi}||}, ~Y_{Bi} = Z_{Bi} \times X_{Bi},
    \end{align} \label{eq:desired_frame}
\end{subequations}
where $\psi_{di}$ is the desired yaw angle, $Y_{Ai}$ is the y-axis of an intermediate coordinate frame used to calculate the desired UAV body frame, represented by the axes $X_{Bi}- Y_{Bi}- Z_{Bi}$ for the $i^{th}$ UAV. Further, the orientation and angular velocity errors, $e_{qi}, \dot{e}_{qi} \forall i \in \lbrace 1,2,\cdots, N \rbrace$ is calculated as,
\begin{align}
    e_q &= \frac{1}{2}(R^T_{i}R_{di} - R^T_{di}R_{i})^v, ~
    \dot{e}_q = R^T_i R_{di}\dot{q}_{di} - \dot{q}_{i},
\end{align}
where $R_{di}$ is the desired rotation matrix obtained from the desired body-frame, coordinate axes in eq. \eqref{eq:desired_frame}, $(.)^v$ is the vee-map, which converts elements from $SO(3)$ to $\mathbb{R}^{3}$. The attitude tracking controller is designed as,
\begin{align}
    \tau_{qi} &= K_{1i}e_{qi} + K_{2i}\dot{e}_{qi} -J_i(R^T_i R_{di}\ddot{q}_{di} -  \dot{q}_iR^T_i R_{di}\dot{q}_{di}) \nonumber \\
    & \quad - \dot{q}_i \times J_i \dot{q}_i
\end{align}
where $K_{1i}, K_{2i} \in \mathbb{R}^{3 \times 3}$ are positive definite gain matrices.

\section{Validiy of CBFs and design of a CBF Filter} \label{sec:cbf_filter}
In this section, we prove that the LCBF presented in Section \ref{sec:prob_form} is a valid time-varying control barrier function as defined in Definition \ref{Def:TV-CBF} and SCBF is a valid CBF. It is to be noted that we validate the CBFs only in the sense of forward invariance, as the aim is to ensure that the UAVs perform the landing starting from a safe region. Then we present the design of a CBF filter that takes a nominal error-correcting controller as input and yields UAV inputs that ensure that the CBF safety constraints are met. The validity of the CBFs for $p_i(t) \in \mathcal{L}_{li} = \mathbb{R}^{3}$ is shown with respect to the following first-order equation that captures the position kinematics of the $N$ UAVs,
\begin{align}
    \mathbf{\dot{p}}(t) &= {\mathbf{u}}, \quad \mathbf{p} \in \mathcal{\Tilde{L}} = \mathbb{R}^{3N} \label{eq:fisrt_order_kinematics}
\end{align}
under the assumption that an inner-loop velocity controller can faithfully track any desired velocity $\mathbf{u}$ in an admissible control input set $\mathbb{U} \subset \mathbb{R}^{3N}$. 

Next, we explicitly define the safe sets involved in the problem of multi-UAV landing, as zero super-level sets of the continuously differentiable CBF functions defined in 
Sections \ref{subsec:landing_cons}- \ref{subsec:collision_cons}.

\subsection{Landing CBF (LCBF)} \label{subsec:lcbf}
The following CBF is derived from Eq. \eqref{eq:lcbf_constraint}, considering the safety and precision constraints
\begin{align}
    h_{li}(p_i,t) &= e_{zi} - \beta_i \alpha_i d_i(t) \exp(- \alpha_i d_i(t)) \label{eq:lcbf}.
\end{align}
If the position of the $i^{th}$ UAV, $p_i \in \mathcal{L}_{li}$ remains in the zero super level set, $\mathcal{V}_{li} = \lbrace p_i \in \mathcal{L}_{li} ~ | ~ h_{li}(p_i,t) \geq 0 \rbrace$ for all $t \geq 0$, then safety is guaranteed when the UAV lands on the UGV. The functions $h_{li}$ are continuously differentiable functions except at $\mathcal{R}_i=\{({x_i}, {y_i}, {z_i}) | e_{xi} = e_{yi} = 0\}$. The functions $h_{li}$ are verified to be time-varying control barrier functions on their respective sets $\mathcal{D}_{li} = \lbrace p_i \in \mathcal{L}_{li} ~ | h_{li}(p_i, t) \in (-b_{li}, \infty), ~ b_{li} > 0  \rbrace \supset \mathcal{V}_{li}$, by deducing the proof for the following lemma.
\begin{lemma}
\label{le:forward_invariance}
    The constraint functions $h_{li}$ in \eqref{eq:lcbf_constraint} are valid time-varying control barrier functions in $\mathcal{L}_{li} \setminus \mathcal{R}_i$ for the dynamics \eqref{eq:fisrt_order_kinematics} and $\kappa_{li}(p_i,t)$, the control input sets are non-empty $\forall p_i \in \mathcal{V}_{li}(t)$. Therefore, the Lipschitz continuous controller $u_i(p_i,t)$ in $\kappa_{li}(p_i,t)$ renders the sets $\mathcal{V}_{li}$ forward invariant in $\mathcal{D}_{li}(t)$.
\end{lemma}
\begin{proof}
    The $h_{li}$ are validated as time-varying CBFs, if there exists the extended extended class-$\mathcal{K}_\infty$ functions $\omega_{li}$ such that
    \begin{align}
        & \sup_{u_i \in \mathcal{U}} \left \lbrace \frac{\partial h_{li}}{\partial p_i} u_i \right \rbrace \geq -\rho_{li} \omega_{li}(h_{li}) - \frac{\partial h_{li}}{\partial t} , \label{eq:lcbf_tv_constraint} \\
        \text{where}  \quad &  \frac{\partial h_{li}}{\partial p_i} = \begin{bmatrix}
           { \frac{\alpha_i \beta_i e_{xi} \exp(-\alpha_i d_i)(\alpha_i d_i - 1)}{d_i}} \\
            {\frac{\alpha_i \beta_i e_{yi} \exp(-\alpha_i d_i)(\alpha_i d_i - 1)}{d_i}} \\
            1
            \end{bmatrix}, \\\frac{\partial h_{li}}{\partial t} &= { \dot{z}_{di} - \frac{(\alpha_i d_i - 1)(e_{xi} \dot{x}_{di}  + e_{yi} \dot{y}_{di})\exp(-\alpha_i d_i)}{d_i}   }         
    \end{align}
    are feasible for $p_i \in \mathcal{D}_{li}$, where $\rho_{li} \in \mathbb{R}$ are positive constants, $u_i$ is the velocity control input of the $i^{th}$ UAV, which is confined by the admissible control input set, $\mathcal{U} = \begin{bmatrix}
        -\sigma & \sigma
    \end{bmatrix} \times \begin{bmatrix}
        -\sigma & \sigma
    \end{bmatrix} \times \begin{bmatrix}
        -\sigma & \sigma
    \end{bmatrix} \subset \mathbb{R}^{3} $. The inequality in the set definition
    \begin{align}
        \kappa_{li}(p_i,t) &= \left \lbrace u_i \in \mathcal{U} ~| ~ \frac{\partial h_{li}}{\partial p_i} u_i \geq -\rho_{li} \omega_{li}(h_{li}) - \frac{\partial h_{li}}{\partial t}  \right 
        \rbrace \label{eq:lcbf_ineq}
    \end{align}
    defines a hyper-plane in $\mathbb{R}^{3}$. When the UAV is in the safe region, ($h_{li}(p_i, t) \geq 0$), $\kappa_{li}(p_i,t)$ contains a ball of nonzero volume centered at the origin of $\mathbb{R}^3$ (i.e., the control input sets are non-empty $\forall p_i \in \mathcal{V}_{li}$), since the last term $\frac{\partial h_{li}}{\partial t}$ in \eqref{eq:lcbf_tv_constraint} is finitely small owing to assumption \ref{as:velocity}. Therefore, the forward invariance of the set $\mathcal{V}_{li}$ is rendered by any Lipschitz continuous $u_i(p_i, t) \in \kappa_{li}$.


    
\end{proof}
\begin{remark}
    Proving the validity of the asymptotic stability aspect of the proposed CBFs is challenging in the presence of input constraints and Lemma \ref{le:forward_invariance} caters only to forward invariance. However, the CBFs are defined in the regions outside the associated safe zones and offer a degree of robustness to forward invariance when initialized with small negative values (as will be demonstrated in the simulation).
    
    The discontinuity in LCBF $h_{li}$ at $e_{xi} = e_{yi}=0$, which forms a vertical line along the Z-axis of the UGV, is handled by directly applying the nominal controller as this line marks the region for initiating the  UAV's vertical descent.
\end{remark}
\subsection{Spherical CBF (SCBF)} \label{subsec:scbf}
The constraints ensuring collision avoidance between the $i^{th}$ and the $j^{th}$ UAVs are captured by the following spherical CBF (SCBF) derived from the constraint \eqref{eq:scbf_constraint},
\begin{align}
        h_{sij}(\widehat{p_{ij}},t) &= || p_i(t) - p_{j}(t) ||^2 - (s_i + s_j)^2. \label{eq:scbf}
\end{align}

If the distance between the $i^{th}$ UAV and the $j^{th}$ UAV is larger than $s_i + s_j$, i.e., if $\widehat{p}_{ij} \triangleq \begin{bmatrix}
     p_i^T & p_j^T
\end{bmatrix} \in \mathcal{L}_{sij} = \mathbb{R}^6$ remains in the zero super level set, $\mathcal{V}_{sij} = \lbrace \widehat{p}_{ij} \in \mathcal{L}_{sij} ~| ~ h_{sij} \in (-b_{sij}, \infty), ~ b_{sij} > 0 \rbrace \,\forall\, t \geq 0$, then the UAVs $i,j$ do not enter into each others unsafe zones parameterized by the radii $s_i$ and $s_j$. Note that unlike LCBF, SCBF is not time-varying as \eqref{eq:scbf} has no explicit dependence on time, that is $\frac{\partial h_{sij}}{\partial t}\equiv 0$. The functions $h_{sij}$ are continuously differentiable and the proof that they are valid control barrier functions, on the sets $\mathcal{D}_{sij} = \lbrace \widehat{p}_{ij} \in \mathcal{L}_{sij} ~| ~ h_{sij} \in (-b_{sij}, \infty), ~ b_{sij} > 0 \rbrace \subset \mathcal{V}_{sij}$ can be deduced following similar arguments as in the proof of Lemma \ref{le:forward_invariance}. 
Therefore, there exists the constants $\rho_{sij} > 0$ and extended class-{$\mathcal{K}_\infty$} functions $\omega_{sij}$ such that
\begin{align}
    & \sup_{\widehat{u}_{ij} \in \mathcal{\widehat{U}}} \left \lbrace \frac{\partial h_{sij}}{\partial \widehat{p}_{ij}} \widehat{u}_{ij} \right \rbrace \geq -\rho_{sij} \omega_{sij}(h_{sij}) 
    , \label{eq:scbf_tv_constraint} \\
        \text{where} \nonumber \\
       \frac{\partial h_{sij}}{\partial p_i} &= \begin{bmatrix}
            2(x_i - x_j) \\
            2(y_i - y_j) \\
            2(z_i - z_j)
        \end{bmatrix}^T, ~ \frac{\partial h_{sij}}{\partial p_j} = -\frac{\partial h_{sij}}{\partial p_i}, ~ \frac{\partial h_{sij}}{\partial t} = 0,
\end{align}
are feasible for $\widehat{p}_{ij} \in \mathcal{D}_{sij} $, where $\widehat{u}_{ij} \triangleq \begin{bmatrix}
     u_i^T & u_j^T
\end{bmatrix}^T$ is in the admissible control input set, $\mathcal{\widehat{U}} \subset \mathbb{R}^{6}$. The input set that ensures the forward invariance of the set $\mathcal{V}_{sij}$ is defined as,
\begin{align}
    \kappa_{sij} = \left \lbrace \widehat{u}_{ij} \in \widehat{\mathcal{U}} ~| ~ \frac{\partial h_{sij}}{\partial \widehat{p}_{ij}} \widehat{u}_{ij} \geq -\rho_{sij} \omega_{sij}(h_{sij})  \right \rbrace.
\end{align}

\subsection{Quadratic programs for safety guarantees}
The CBF filter takes in a nominal pseudo-velocity control $\mathbf{u}'(\mathbf{p},t)$ from the position controller in \eqref{eq:position_control}, and guaranteed safe outputs the closest velocity control input, $\mathbf{u^*}(\mathbf{p},t)$. The nominal controller is designed such that the UAVs asymptotically reach the landing locations on the respective UGV. The CBF filter is formulated as the following Quadratic program
\begin{align}
    \mathbf{u^*} = & \argmin_{\mathbf{u} \in \mathbb{U}} ||\mathbf{u} - \mathbf{u}'(\mathbf{p},t) ||^2, \nonumber \\
     & \text{s.t.: } \mathbf{A}\mathbf{u} \geq - \mathbf{b} \label{eq:cbf_form}
\end{align}
where the linear constraint in \eqref{eq:cbf_form} is built from the CBF constraints ($\frac{\partial h(p,t)}{\partial p} \Leftrightarrow \mathbf{A}$ with $f(p) = 0, ~g(p) = \mathbf{I}$ and $\mathbf{b} \Leftrightarrow  \omega (h(p,t)) + \frac{\partial h(p,t)}{\partial t} $ cf. \eqref{eq:cbf_def}) associated with the CBFs defined so far in the article. In the linear constraint \eqref{eq:cbf_form}, the matrices $\mathbf{A} \in \mathbb{R}^{Q \times 3N}, \mathbf{b} \in \mathbb{R}^Q$, where $N$ is the number of UAVs and $Q = \frac{N(N+1)}{2} $. The matrix $\mathbf{A}$ is a sparse matrix containing the partial derivatives of the CBFs. Let $A[i,j] \in \mathbb{R}^{3}$ be a set of row matrices such that $A[i,j] \triangleq \mathbf{A}[i, (3(j-1)+1):3j]$. Now, the elements of $\mathbf{A}, \mathbf{b}$ are given by,
\begin{align}
    A[i,i] &= \frac{\partial h_{li}^T}{\partial p_i}, ~ 1 \leq i \leq N, \label{eq:A_1}\\
    {A}[\nu,i] &= \frac{\partial h_{sij}^T}{\partial p_i}, ~ {A}[\nu,j] = \frac{\partial h_{sij}^T}{\partial p_j} , \quad 1 \leq i \leq j \leq N, \label{eq:A_2} \\
    \text{where } \nu &= N + i + j - 2 + \sum_{k}^{i-1} \left ( N - k - 2 \right ), \nonumber \\
    \mathbf{b}[i] &=  {\rho_{li} }h_{li} + \frac{\partial h_{li}}{\partial t} , \quad 1 \leq i \leq N, \label{eq:b_1}\\
    \mathbf{b}[\nu] &=  {\rho_{sij}} h_{sij} + \frac{\partial h_{sij}}{\partial t}, \quad 1 \leq i \leq j \leq N. \label{eq:b_2}
\end{align}

The constraint \eqref{eq:cbf_form} is linear in $\mathbf{u}$ for a given $(\mathbf{p}(t), \mathbf{p_d}(t), t)$ with the matrix $\mathbf{A}$ as defined in \eqref{eq:A_1}-\eqref{eq:A_2}. At any time $t$, the classical quadratic program \eqref{eq:cbf_form} can be solved efficiently at very high rates.
\subsection{Input sharing among multiple CBFs} \label{sec:inp_sha}
Although the functions $h_{li}$ and $h_{sij}$ are proven to individually render the sets $\mathcal{V}_{li}$ and $\mathcal{V}_{sij}$ invariant respectively, their efficacy when enforced together has to be further analyzed since they share the same control inputs $\mathbf{u}$ in the quadratic program \eqref{eq:cbf_form}. At this stage, let us recall the concept of control-sharing time-varying CBFs in multi-input sharing first-order barrier functions from \cite{saradagi2022safe}, but in it's time-varying form.
\begin{definition}[Control Sharing Time-varying CBFs] \label{def:control_sharing}
    The time-varying CBFs $h_i(p,t)$ defined for a control affine dynamical system are considered to be control sharing time-varying CBFs, if there exists a control input $u \in \mathcal{U}$ such that for any $p \in \mathcal{P}$
    \begin{align}
        \dot{h}_i(p, t) \geq - \rho_i \omega_i(h_i(p, t)), ~ \forall i \in \lbrace 1, \cdots, N \rbrace \label{eq:control_sharing_condition}
    \end{align}
\end{definition}
In this article, the CBF functions in \eqref{eq:lcbf} and \eqref{eq:scbf} in the state space of $\mathbf{p} \in \mathbb{P} = \mathbb{R}^{3N}$ share the control inputs $\mathbf{u} \in \mathbb{U} = \mathbb{R}^{3N} $. 
Though control sharing is extremely challenging and infeasible in many cases, it is slightly less challenging in our case, mainly because: 1) The constraint matrix $\mathbf{A}$ becomes sparser with the increase in the number of UAVs $N$, as not more than six values are non-zero in each row for a row length of $3N$; 2) The LCBFs $h_{li} ~\forall i \in \lbrace 1, \cdots, N \rbrace$ are independent of each other and constrain the states of only one UAV each. 3) The LCBFs have no effect on the nominal control when the UAVs are above a certain altitude relative to their corresponding UGVs.

Since the CBFs are combined in the constraint matrix $\mathbf{A}$, they act on the state space of $\mathbf{p} \in \mathbb{P}$. Therefore, let us define the appended form of the functions $h_{li}, h_{sij}$ as $\Tilde{h}_{li}, \Tilde{h}_{sij}$, and appended form of the sets $\mathcal{{V}}_{li}, \mathcal{{V}}_{sij}$ as $\mathcal{\Tilde{V}}_{li}, \mathcal{\Tilde{V}}_{sij}$ such that, $\mathcal{\Tilde{V}}_{li} = \lbrace \mathbf{p} \in \mathcal{\Tilde{L}} ~ | ~ \Tilde{h}_{li}(\mathbf{p},t) \geq 0 \rbrace$, $\mathcal{\Tilde{V}}_{sij} = \lbrace \mathbf{p} \in \mathcal{\Tilde{L}} ~ | ~ \Tilde{h}_{sij}(\mathbf{p},t) \geq 0 \rbrace$, for all $t\geq0$. Now, we discuss the conditions necessary for the control-sharing property of the CBFs to render the forward invariance of the intersection of the sets $\mathcal{H} = \mathcal{\Tilde{V}}_{li} \cap \mathcal{\Tilde{V}}_{sij}$.
Now, the appended form of control sets $\kappa_{li}, \kappa_{sij}$ is given by $\Tilde{\kappa}_{li}, \Tilde{\kappa}_{sij}$, such that
\begin{align}
    \Tilde{\kappa}_{n}(\mathbf{p},t) &= \left \lbrace \mathbf{u} \in \mathbb{U} ~| ~ \frac{\partial \Tilde{h}_{n}}{\partial \mathbf{p}} \mathbf{u} \geq -\Tilde{\rho}_{n} \Tilde{\omega}_{n}(\Tilde{h}_{n}) - \frac{\partial \Tilde{h}_{n}}{\partial t}  \right \rbrace,  \nonumber
\end{align}
where $\Tilde{\rho}_{n} > 0, ~\Tilde{\omega}_{n}$ is a class-$\mathcal{K}_\infty$ function $\forall \, n \in \lbrace li, sij\rbrace$.
\begin{proposition} \label{prpsn:control_sharing}
    The intersection of the sets $\Tilde{\kappa}_{li}, \Tilde{\kappa}_{sij}$ is non-empty for the functions $\Tilde{h}_{li}, \Tilde{h}_{sij}$. The Lipschitz continuous controller $\mathbf{u}(\mathbf{p}, t) \in \Tilde{\kappa}_{li} \cap \Tilde{\kappa}_{sij} $ for the dynamic system \eqref{eq:fisrt_order_kinematics} renders the forward invariance of the set $\mathcal{H}$.
\end{proposition}
\begin{proof}
    The conditions for the SCBFs in \eqref{eq:scbf_tv_constraint} define half spaces in the input space $H_{sij}(\mathbf{p},t) \in \mathbb{R}^{3N}$. Since the partial derivative of the SCBFs with respect to time is zero (SCBFs are time-invariant), the half-spaces formed by the SCBFs contain the nonzero volumed ball centered at the origin within them at any time, when $\Tilde{h}_{sij}(t) \geq 0$. Also, since $\Tilde{h}_{sij} \geq 0$ before take-off and after landing (cf. assumption \ref{as:position}), a non-empty set $\boldsymbol{\kappa_s}(\mathbf{p},t)$ is formed by the intersection of all $\Tilde{\kappa}_{sij}(\mathbf{p},t)$.
    
    Further, the half spaces $H_{li}(\mathbf{p},t) \in \mathbb{R}^{3N}$ are independent of each other. Hence, a non-empty set $\boldsymbol{\kappa_l}(\mathbf{p},t)$ containing a ball of nonzero radius centered at the origin is formed in their intersection. Therefore, the intersection of the permissible control sets
    $\boldsymbol{\kappa_s}(\mathbf{p},t) \cap \boldsymbol{\kappa_l}(\mathbf{p},t)$ is non-empty when their relative velocities are bounded owing to assumption \ref{as:velocity}. Thus, the forward invariance of the set $\mathcal{H}$ under any Lipschitz continuous controller $\mathbf{u}^*(\mathbf{p},t) \in \boldsymbol{\kappa_s}(\mathbf{p},t) \cap \boldsymbol{\kappa_l}(\mathbf{p},t) $ is concluded by the virtue of Proposition \ref{prpsn:asymptotic}.
\end{proof}
\begin{remark}[Parameter Tuning]
    The finite-time landing depends on the gains of the nominal controller $\mathbf{K_p}, \mathbf{K_v}$. Higher values of $\mathbf{K_p}, \mathbf{K_v}$ result in faster convergence. However, large gain values also demand large control inputs, which is undesirable.
\end{remark}

\section{Simulated Results} \label{sec:sim_res}
\begin{figure}[!h]
	\centering
	\includegraphics[width=0.48\textwidth]{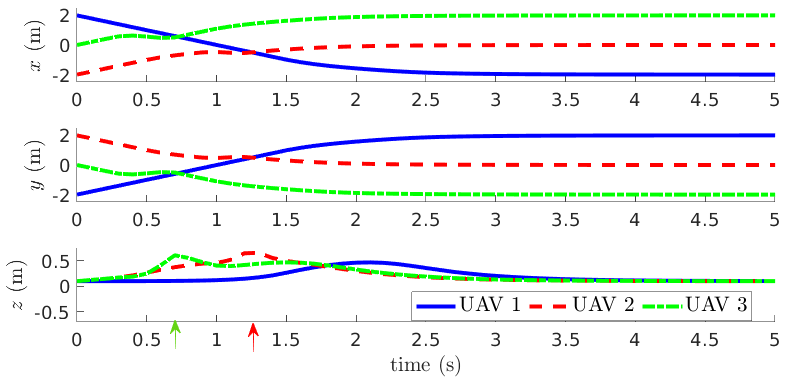}
	\caption{{Trajectories taken by the UAVs to reach their respective UGVs in Scenario 1. At times $0.7$ s, $1.3$ s the altitudes of UAV 3 and UAV 2 respectively, rise to avoid collision with UAV 1.}}\label{fig:sce_1_2d}
\end{figure}
\begin{figure}[!h]
	\centering
	\includegraphics[width=0.48\textwidth]{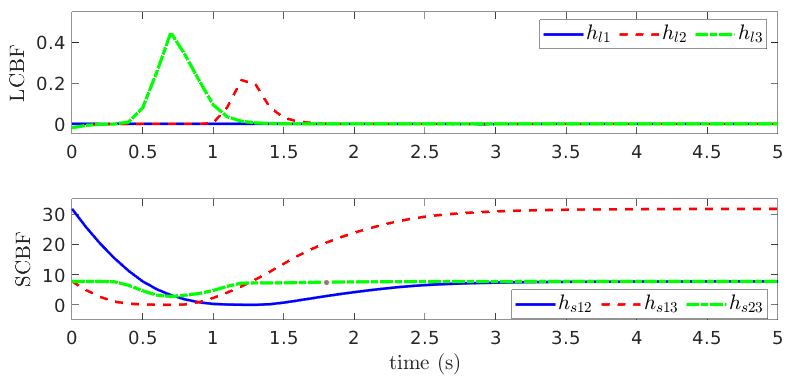}
	\caption{{Values of the CBFs over time in Scenario 1.}}\label{fig:sce_1_h}
\end{figure}
\begin{figure}[!h]
	\centering
	\includegraphics[width=0.48\textwidth]{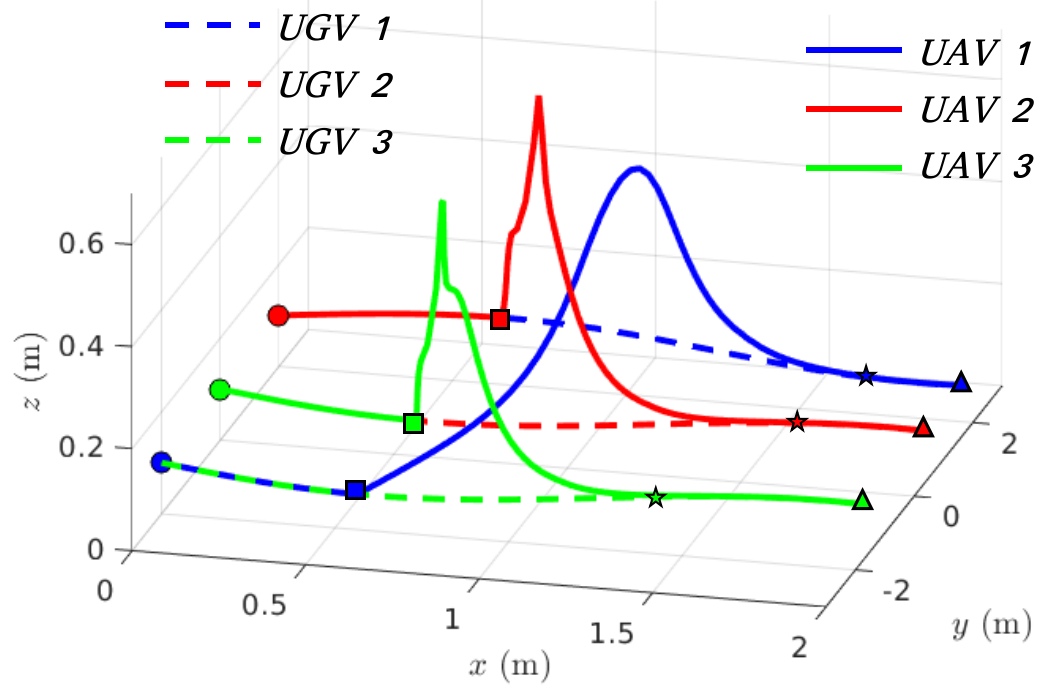}
	\caption{{Three-dimensional plot of the trajectories taken by the UAVs and UGVs in Scenario 2. The circles mark the start of the trajectories; the squares mark the points where the controllers are turned ON; the landing points are represented by stars, and triangles mark the ends of the trajectories.}}\label{fig:sce_2_traj_3d}
\end{figure}
\begin{figure}[!h]
	\centering
	\includegraphics[width=0.48\textwidth]{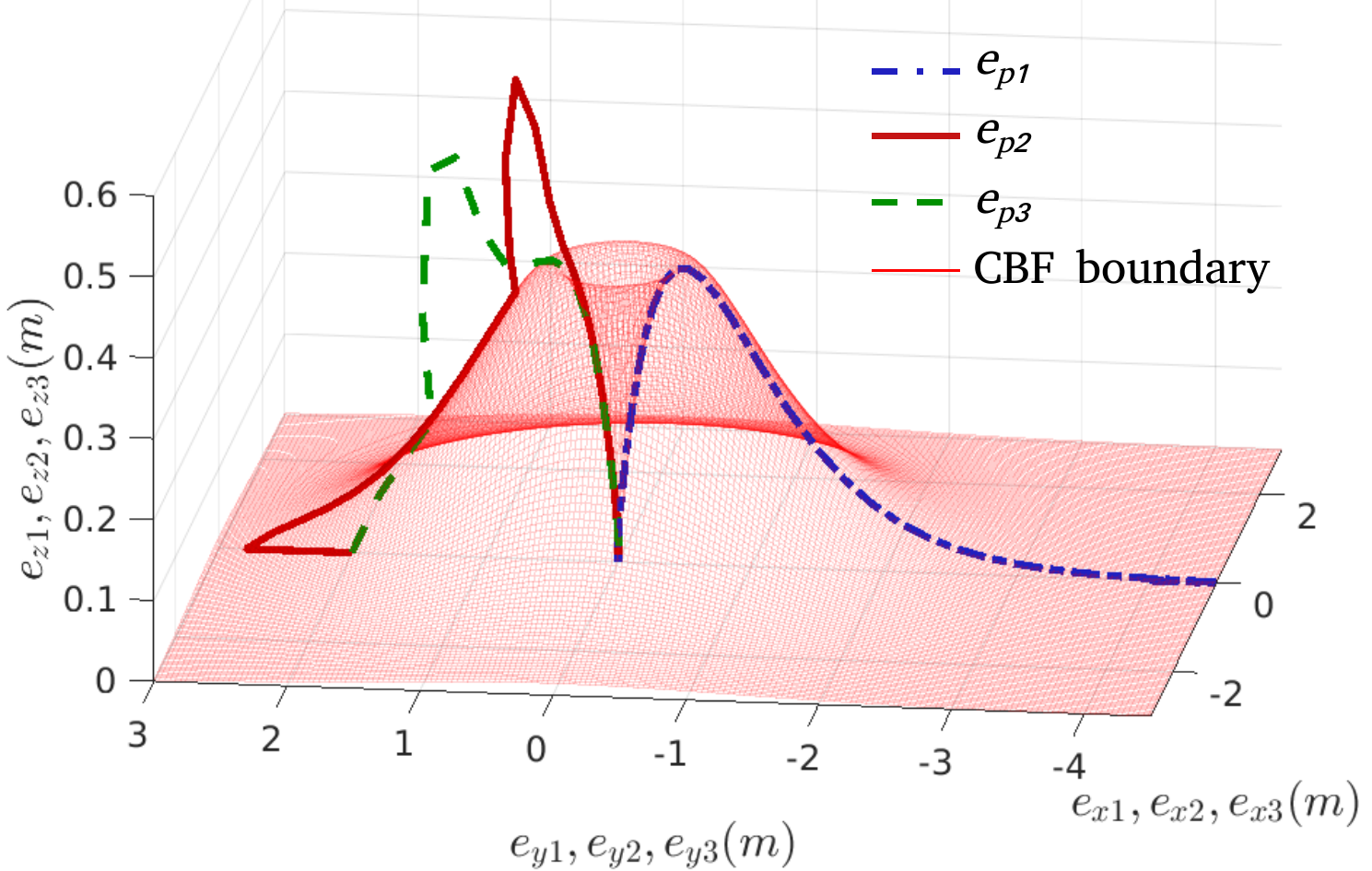}
	\caption{{Three-dimensional representation of the LCBFs in terms of $e_{p1}, e_{p2}, e_{p3}$ in Scenario 2. The CBF boundaries are static in these reference frames. The trajectories start from safe regions away from the origin and converge to the origin without crossing the boundary layer.}}\label{fig:lcbf_3d}
\end{figure}
\begin{figure}[!h]
	\centering
	\includegraphics[width=0.48\textwidth]{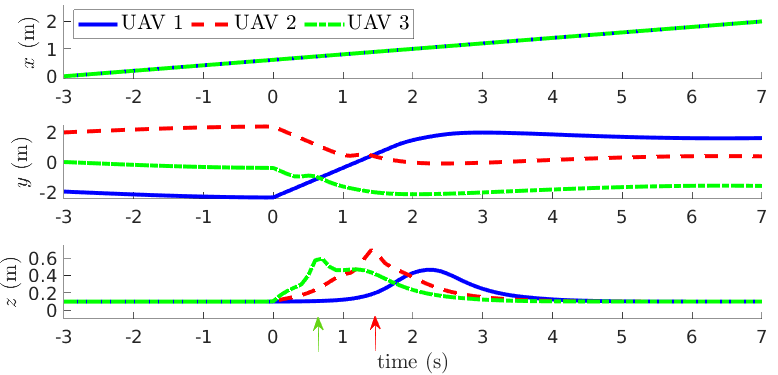}
	\caption{{Trajectories taken by the UAVs to reach their respective UGVs in Scenario 2. At times $0.7$ s, $1.3$ s the altitudes of UAV 3 and UAV 2 respectively, rise to avoid collision with UAV 1.}}\label{fig:sce_2_2d}
\end{figure}
\begin{figure}[!h]
	\centering
	\includegraphics[width=0.48\textwidth]{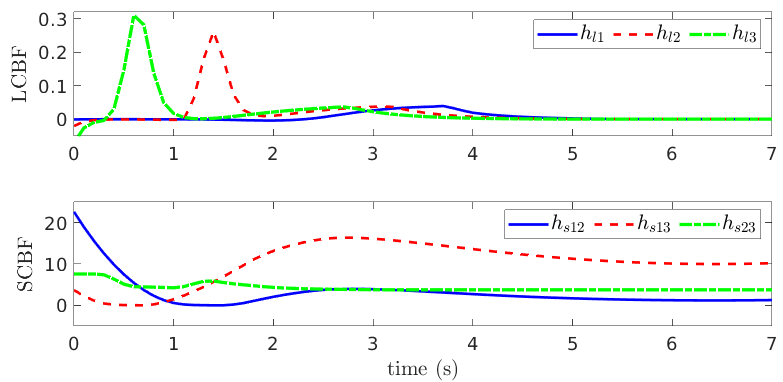}
	\caption{{Value of the CBFs over time, where the time at which the controller is turned ON is considered to be $t=0$ in Scenario 2.}}\label{fig:sce_2_h}
\end{figure}
The performance of the controller is validated in a simulated environment in MATLAB. The dynamics of the UAVs are modeled using the Astec-Pelican quadrotor model, with a mass of $1$ kg each. The control parameters used in the simulation are: $\mathbf{K_p} = \mathbf{K_v} = 2\mathbf{I} \in \mathbb{R}^{3N \times 3N}$, $K_{1i} = K_{2i} = 0.5 \mathbf{I} \in \mathbb{R}^{3 \times 3}$, $\alpha_i = 2.0$, $\beta_i = 1.0$, $a_k = 1.0$, $b_k = 3.0$, $s_i = s_j = 0.25$m, $m_k = 0.5$m, $\rho_{li} = \rho_{sij} = 10.0$, $\sigma = 2$, for all $i \in \lbrace 1, \cdots, N \rbrace$, $i < j \leq N$. The simulated setup has three UAV\textendash UGV pairs, such that UAV 1 is carried by UGV 3 and UAVs 2 and 3  are carried by UGVs 1 and 2, respectively. The controller's performance is tested in two scenarios with static and moving UGVs respectively. 

In Scenario 1, the controller is turned ON immediately when the simulation begins with the initial conditions $p_1(0) = p_{d3}(0) = \begin{bmatrix}
    2.0 & -2.0 & 0.1
\end{bmatrix}^{T}$, $p_2(0) = p_{d1}(0) = \begin{bmatrix}
    -2.0 & 2.0 & 0.1
\end{bmatrix}^{T}$, $p_3(0) = p_{d2}(0) = \begin{bmatrix}
    0.0 & 0.0 & 0.1
\end{bmatrix}^{T}$. 
In scenario 2, UGVs move along a predefined path while carrying the UAVs. The controllers are turned ON at 3s after the start of the simulation. The time at which the UAVs' controllers are turned ON is considered to be $t=0$. The initial conditions for Scenario 2, at time $t = -3$s are given by, $p_1(-3) = p_{d3}(-3) = \begin{bmatrix}
    0.0 & -2.0 & 0.1
\end{bmatrix}^{T}$, $p_2(-3) = p_{d1}(-3) = \begin{bmatrix}
    0.0 & 2.0 & 0.1
\end{bmatrix}^{T}$, $p_3(-3) = p_{d2}(-3) = \begin{bmatrix}
    0.0 & 0.0 & 0.1
\end{bmatrix}^{T}$, $\dot{p}_{d1}(t) = \begin{bmatrix}
    0.2 & 0.2\cos(0.5t) & 0.0
\end{bmatrix}^{T}$, $\dot{p}_{d2}(t) = \dot{p}_{d3}(t) = \begin{bmatrix}
    0.2 & -0.2\cos(0.5t) & 0.0
\end{bmatrix}^{T}$. 

The results of Scenarios 1 and 2 are highlighted in Fig. \ref{fig:sce_1_2d}- \ref{fig:sce_1_h} and Fig. \ref{fig:sce_2_traj_3d}-\ref{fig:sce_2_h} respectively. Fig. \ref{fig:sce_1_2d} shows the trajectories of the UAVs. It is evident from Fig. \ref{fig:sce_1_2d} that the UAVs gain an altitude when they get close to their corresponding UGVs and complete the landing with a vertical descent. Further, they maintain a minimum altitude in all places outside the proximity of their corresponding UGVs. The same is reflected in the LCBF plot in Fig. \ref{fig:sce_1_h}, where the $h_{li}$ values turn positive from their small negative initial conditions and remain non-negative thereafter, thus proving the forward invariance of the safe set. It is to be noted that the $h_{li}$ values spike at the instances, in which the UAVs get close to each other in the horizontal plane. In these instances, their corresponding SCBFs get close to zero, resulting in the controller providing a control input to push UAV 3 and UAV 2 to a higher altitude (cf. Fig. \ref{fig:sce_1_2d} at time 0.7 s and 1.3 s) above UAV 1 to maintain a safe Euclidean distance of 0.5 m.

When the UGVs are moving in Scenario 2, the trajectories plotted in Fig. \ref{fig:sce_2_traj_3d} show that when the controller is turned ON, the UAVs move from their carrier UGVs to land on their respective UGVs while also maintaining a safe distance between other UAVs. In Fig. \ref{fig:lcbf_3d}, the boundaries of the landing CBFs $h_{li}$ are overlaid on each other and represented in the frames of $e_{xi}-e_{yi}-e_{zi}$, in which the LCBFs are stationary. The position error trajectories provide a perspective of how the UAVs perform the vertical descent and stay above the LCBF throughout its spread. Similar to Scenario 1, the values of $h_{li}$ start from a small negative value and converge to zero when the controller is started. It is to be noted that the LCBF plots are shown from the time when the controller is turned ON, while the position trajectories begin from $-3$ s to show the movement of the UAVs in the overall Scenario. Even in this scenario, when the UAVs get close to each other horizontally, the altitudes of UAV 2 and UAV 3 increase as a result of SCBFs, which is reflected in the position trajectories in Fig. \ref{fig:sce_2_2d} and the values of SCBFs in Fig. \ref{fig:sce_2_h}. 

From these observations, it is inferred that the forward invariance of the CBFs is rendered even when the control inputs are shared between multiple CBFs. Moreover, the LCBFs asymptotically converge to the safe set when initiated close to the boundary and outside the safe set.

\section{Conclusions} \label{sec:concl}
In this article, the landing problem of multiple UAVs on moving UGVs is formulated as a multiple CBF control problem. The constraints are formulated systematically, and two classes of CBFs are designed to accommodate the constraints. A nominal controller is designed to track the landing position, and its control output is filtered using the CBFs. The control-sharing property of the CBFs is validated in terms of forward invariance. The effectiveness of the proposed control architecture is validated using a MATLAB simulation, and the analysis of the results is presented.

\bibliographystyle{IEEEtran}
\bibliography{ref}
\end{document}